\documentclass[11pt,letterpaper]{article}
\usepackage{emnlp2017}
\usepackage{times}
\usepackage{latexsym}

\usepackage{graphicx}
\usepackage{booktabs}
\usepackage{tabularx}
\usepackage{amsmath}
\usepackage{color}
\usepackage{pgfplots}

\emnlpfinalcopy

\def\emnlppaperid{***}

\title{Artificial Error Generation \\with Machine Translation and Syntactic Patterns}

\author{Marek Rei \hspace{1cm} Mariano Felice \hspace{1cm} Zheng Yuan \hspace{1cm} Ted Briscoe\\
	    The ALTA Institute\\
	    Computer Laboratory\\
	    University of Cambridge\\
        United Kingdom\\
	    {\tt firstname.lastname@cl.cam.ac.uk} \\}

\date{}

\begin{document}
\maketitle

\begin{abstract}
Shortage of available training data is holding back progress in the area of automated error detection.
This paper investigates two alternative methods for artificially generating writing errors, in order to create additional resources.
We propose treating error generation as a machine translation task, where grammatically correct text is translated to contain errors.
In addition, we explore a system for extracting textual patterns from an annotated corpus, which can then be used to insert errors into grammatically correct sentences.
Our experiments show that the inclusion of artificially generated errors significantly improves error detection accuracy on both FCE and CoNLL 2014 datasets.
\end{abstract}

\section{Introduction}
Writing errors can occur in many different forms -- from relatively simple punctuation and determiner errors, to mistakes including word tense and form, incorrect collocations and erroneous idioms. %pronouns
Automatically identifying all of these errors is a challenging task, especially as the amount of available annotated data is very limited. 
\newcite{Rei2016} showed that while some error detection algorithms perform better than others, it is additional training data that has the biggest impact on improving performance.

Being able to generate realistic artificial data would allow for any grammatically correct text to be transformed into annotated examples containing writing errors, producing large amounts of additional training examples.
Supervised error generation systems would also provide an efficient method for anonymising the source corpus -- error statistics from a private corpus can be aggregated and applied to a different target text, obscuring sensitive information in the original examination scripts.
However, the task of creating incorrect data is somewhat more difficult than might initially appear -- naive methods for error generation can create data that does not resemble natural errors, thereby making downstream systems learn misleading or uninformative patterns.

Previous work on artificial error generation (AEG) has focused on specific error types, such as prepositions and determiners \cite{Rozovskaya2010a,Rozovskaya2011a}, or noun number errors \cite{Brockett2006}.
\newcite{Felice2014a} investigated the use of linguistic information when generating artificial data for error correction, but also restricting the approach to only five error types. 
There has been very limited research on generating artificial data for all types, which is important for general-purpose error detection systems.
For example, the error types investigated by \newcite{Felice2014a} cover only 35.74\% of all errors present in the CoNLL 2014 training dataset, providing no additional information for the majority of errors.

In this paper, we investigate two supervised approaches for generating all types of artificial errors.
We propose a framework for generating errors based on statistical machine translation (SMT), training a model to translate from correct into incorrect sentences. In addition, we describe a method for learning error patterns from an annotated corpus and transplanting them into error-free text. We evaluate the effect of introducing artificial data on two error detection benchmarks.
Our results show that each method provides significant improvements over using only the available training set, and a combination of both gives an absolute improvement of 4.3\% in $F_{0.5}$, without requiring any additional annotated data.

\begin{table*}[t]
\setlength\tabcolsep{7pt}
\begin{tabular}{l|l} \hline
\normalsize
Original & We are a well-mixed class with equal numbers of boys and girls, all about 20 years old. \\
%Random & We \textbf{years} a well-mixed class with equal numbers of boys and girls, all about 20 years old. \\
FY14 & We \textbf{am} a well-mixed class with equal numbers of boys and girls, all about 20 years old. \\
PAT & We are a well-mixed class with equal numbers of boys \textbf{an} girls, all about 20 \textbf{year} old. \\
MT & We are a well-mixed class with \textbf{equals} numbers of boys and girls, all about 20 years old. \\

%Original & First of all, I want to highlight that we see an awful lot of mobile phones in our daily life. \\
%Random & First of all, I want to highlight that we see an awful lot of mobile \textbf{that} in our daily life. \\
%Patterns & First of all, I want to highlight \textbf{out} that we see an awful lot of mobile phones in our daily life. \\

%Original & On average, our pupils are 16 years old and they join actively in all the events we organise.  \\ 
%Random & On average, our pupils are 16 years old \textbf{16} and they join actively in all the events we organise. \\
%Patterns & On average, our pupils are 16 \textbf{year} old and they join actively in all the events we organise.  \\ 
%MT & On average, our pupils are 16 years old and they join \textbf{in} actively in all the events we organise. \\ 
\hline
\end{tabular}
\caption{Example artificial errors generated by three systems: the error generation method by \newcite{Felice2014a} (FY14), our pattern-based method covering all error types (PAT), and the machine translation approach to artificial error generation (MT).}
\label{tab:examples}
\end{table*}

\hspace{2cm}

\section{Error Generation Methods}
\label{sec:methods}

We investigate two alternative methods for AEG. The models receive grammatically correct text as input and modify certain tokens to produce incorrect sequences. The alternative versions of each sentence are aligned using Levenshtein distance, allowing us to identify specific words that need to be marked as errors. While these alignments are not always perfect, we found them to be sufficient for practical purposes, since alternative alignments of similar sentences often result in the same binary labeling.
Future work could explore more advanced alignment methods, such as proposed by \newcite{felice-bryant-briscoe}.

In Section \ref{sec:evaluation}, this automatically labeled data is then used for training error detection models.

\subsection{Machine Translation}

We treat AEG as a translation task -- given a correct sentence as input, the system would learn to translate it to contain likely errors, based on a training corpus of parallel data.
Existing SMT approaches are already optimised for identifying context patterns that correspond to specific output sequences, which is also required for generating human-like errors.
The reverse of this idea, translating from incorrect to correct sentences, has been shown to work well for error correction tasks \cite{Brockett2006,Ng2013a}, and round-trip translation has also been shown to be promising for correcting grammatical errors \cite{madnani-tetreault-chodorow}.

Following previous work \cite{Brockett2006,Yuan-Felice:2013}, we build a phrase-based SMT error generation system. 
During training, error-corrected sentences in the training data are treated as the source, and the original sentences written by language learners as the target. Pialign~\cite{Pialign} is used to create a phrase translation table directly from model probabilities. 
In addition to default features, we add character-level Levenshtein distance to each mapping in the phrase table, as proposed by \newcite{Felice:2014-CoNLL}. 
Decoding is performed using Moses~\cite{Moses} and the language model used during decoding is built from the original erroneous sentences in the learner corpus. 
The IRSTLM Toolkit~\cite{IRSTLM} is used for building a 5-gram language model with modified Kneser-Ney smoothing~\cite{Kneser1995}.

\subsection{Pattern Extraction}
\label{sec:pat}

We also describe a method for AEG using patterns over words and part-of-speech (POS) tags, extracting known incorrect sequences from a corpus of annotated corrections.
This approach is based on the best method identified by \newcite{Felice2014a}, using error type distributions; while they covered only 5 error types, we relax this restriction and learn patterns for generating all types of errors.%, which we have extended to cover all error types.

The original and corrected sentences in the corpus are aligned and used to identify short transformation patterns in the form of \textit{(incorrect phrase, correct phrase)}. The length of each pattern is the affected phrase, plus up to one token of context on both sides. If a word form changes between the incorrect and correct text, it is fully saved in the pattern, otherwise the POS tags are used for matching.

For example, the original sentence `\textit{We went shop on Saturday}' and the corrected version `\textit{We went shopping on Saturday}' would produce the following pattern:
%\vspace{3mm}
\begin{center}
\noindent {\small \textit{(VVD shop\_VV0 II, VVD shopping\_VVG II)}}
\end{center}
%\vspace{3mm}

\noindent After collecting statistics from the background corpus, errors can be inserted into error-free text. The learned patterns are now reversed, looking for the correct side of the tuple in the input sentence. 
We only use patterns with frequency $>= 5$, which yields a total of 35,625 patterns from our training data. 
%The number of errors in a sentence is first chosen based on probabilities from the background corpus, and then errors are sampled from the set of applicable patterns.
For each input sentence, we first decide how many errors will be generated (using probabilities from the background corpus) and attempt to create them by sampling from the collection of applicable patterns. This process is repeated until all the required errors have been generated or the sentence is exhausted. During generation, we try to balance the distribution of error types as well as keeping the same proportion of incorrect and correct sentences as in the background corpus \cite{UCAM-CL-TR-895}. The required POS tags were generated with RASP \cite{Briscoe2006}, using the CLAWS2 tagset.%\footnote{\url{http://ucrel.lancs.ac.uk/claws2tags.html}}

\begin{table*}[t]
\setlength\tabcolsep{8pt}
\begin{tabular}{l|ccc|ccc|ccc} \hline
 & \multicolumn{3}{c|}{FCE} & \multicolumn{3}{c|}{CoNLL-14 TEST1} & \multicolumn{3}{c}{CoNLL-14 TEST2} \\ 
 & {\small P} & {\small R} & {\small $F_{0.5}$} & {\small P} & {\small R} & {\small $F_{0.5}$} & {\small P} & {\small R} & {\small $F_{0.5}$} \\\hline
R\&Y (2016) & 46.10 & 28.50 & 41.10 & 15.40 & \textbf{22.80} & 16.40 & 23.60 & \textbf{25.10} & 23.90\\ \hline
Annotation & 53.91 & 26.88 & 44.84 & 16.12 & 18.42 & 16.52 & 25.72 & 20.92 & 24.57\\
Ann+FY14 & 58.77 & 25.55 & 46.54 & 20.48 & 14.41 & 18.88 & 33.25 & 16.67 & 27.72\\
Ann+PAT & \textbf{62.47} & 24.70 & 47.81 & 21.07 & 15.02 & 19.47 & 34.04 & 17.32 & 28.49\\
Ann+MT & 58.38 & \textbf{28.84} & 48.37 & 19.52 & 20.79 & 19.73 & 30.24 & 22.96 & 28.39\\
Ann+PAT+MT & 60.67 & 28.08 & \textbf{49.11} & \textbf{23.28} & 18.01 & \textbf{21.87} & \textbf{35.28} & 19.42 & \textbf{30.13} \\ \hline
\end{tabular}
\caption{Error detection performance when combining manually annotated and artificial training data.}
\label{tab:add}
\end{table*}

\section{Error Detection Model}
\label{sec:model}

We construct a neural sequence labeling model for error detection, following the previous work \cite{Rei2016,Rei2017}. % and \newcite{Lample2016}. 
The model receives a sequence of tokens
% $(w_1, ..., w_T)$ 
as input and outputs a prediction for each position, indicating whether the token is correct or incorrect in the current context.
The tokens are first mapped to a distributed vector space, resulting in a sequence of word embeddings. % $(x_1, ..., x_T)$.
Next, the embeddings are given as input to a bidirectional LSTM \cite{Hochreiter1997}, in order to create context-dependent representations for every token.
The  hidden states from forward- and backward-LSTMs are concatenated for each word position, resulting in representations that are conditioned on the whole sequence. %:
%$$
%\overrightarrow{h_t} = LSTM(x_t, \overrightarrow{h_{t-1}})
%$$
%$$
%\overleftarrow{h_t} = LSTM(x_t, \overleftarrow{h_{t+1}})
%$$
%$$
%h_t = [\overrightarrow{h_t};\overleftarrow{h_t}]
%$$
%This concatenated vector is then passed through an additional feedforward layer, in order to combine the extracted features and map them to a more suitable space.
%:
%$$
%d_t = tanh(W_d h_t)
%$$
%\noindent where $W_d$ is a weight matrix between the layers.
%Finally, a softmax over the two possible labels (\textit{correct} and \textit{incorrect}) is used to output a probability distribution for each token.
%:
%$$
%P(y_t = k | d_t) = \frac{e^{W_{o,k} d_t}}{\sum_{\tilde{k} \in K} e^{W_{o,\tilde{k}} d_t}}
%$$
%\noindent where $P (y_t = k|d_t )$ is the probability of the label of the $t$-th word ($y_t$) being $k$, $K$ is the set of all possible labels, and $W_{o,k}$ is the $k$-th row of output weight matrix $W_o$. 
%The model is optimised by minimising categorical cross-entropy, which is equivalent to minimising the negative log-probability of the correct labels.
This concatenated vector is then passed through an additional feedforward layer, and a softmax over the two possible labels (\textit{correct} and \textit{incorrect}) is used to output a probability distribution for each token.
The model is optimised by minimising categorical cross-entropy with respect to the correct labels.
We use AdaDelta \cite{Zeiler2012} for calculating an adaptive learning rate during training, which accounts for a higher baseline performance compared to previous results.
%:

%$$
%E = - \sum_{t=1}^{T} log(P(y_t| d_t))
%$$

%\begin{table}[t]
%\setlength\tabcolsep{5pt}
%\begin{tabular}{llrrr} \hline
% & &  & \multicolumn{2}{c}{CoNLL-14} \\
%Method & Dataset & FCE & Test-1 & Test-2 \\ \hline
% & SimpleWiki & 12.59 & 9.06 & 9.34 \\
%Patterns & FCE & 10.30 & 3.73 & 4.45 \\
% & EVP & 9.32 & 4.55 & 5.69 \\ \hline
% & SimpleWiki & 23.50 & \textbf{16.35} & 19.09 \\
%MT & FCE & \textbf{34.27} & 16.13 & \textbf{21.40} \\
% & EVP & 27.93 & 16.04 & 18.95 \\ \hline
%\end{tabular}
%\caption{Error detection $F_{0.5}$ score using different sources of artificial training data.}
%\label{tab:main}
%\end{table}

\section{Evaluation}
\label{sec:evaluation}

We trained our error generation models on the public FCE training set \cite{Yannakoudakis2011} and used them to generate additional artificial training data.
Grammatically correct text is needed as the starting point for inserting artificial errors, and we used two different sources: 1) the corrected version of the same FCE training set on which the system is trained (450K tokens), and 2) example sentences extracted from the English Vocabulary Profile (270K tokens).\footnote{http://www.englishprofile.org/wordlists}. While there are other text corpora that could be used (e.g., Wikipedia and news articles), our development experiments showed that keeping the writing style and vocabulary close to the target domain gives better results compared to simply including more data. 

We evaluated our detection models on three benchmarks: the FCE test data (41K tokens) and the two alternative annotations of the CoNLL 2014 Shared Task dataset (30K tokens) \cite{Ng2013a}.
Each artificial error generation system was used to generate 3 different versions of the artificial data, which were then combined with the original annotated dataset and used for training an error detection system.
Table \ref{tab:examples} contains example sentences from the error generation systems, highlighting each of the edits that are marked as errors.
The error detection results can be seen in Table \ref{tab:add}. We use $F_{0.5}$ as the main evaluation measure, which was established as the preferred measure for error correction and detection by the CoNLL-14 shared task \cite{Ng2013a}. $F_{0.5}$ calculates a weighted harmonic mean of precision and recall, which assigns twice as much importance to precision -- this is motivated by practical applications, where accurate predictions from an error detection system are more important compared to coverage. For comparison, we also report the performance of the error detection system by \newcite{Rei2016}, trained using the same FCE dataset.

The results show that error detection performance is substantially improved by making use of artificially generated data, created by any of the described methods.
When comparing the error generation system by \newcite{Felice2014a} (FY14) with our pattern-based (PAT) and machine translation (MT) approaches, we see that the latter methods covering all error types consistently improve performance. 
While the added error types tend to be less frequent and more complicated to capture, the added coverage is indeed beneficial for error detection.
Combining the pattern-based approach with the machine translation system (Ann+PAT+MT) gave the best overall performance on all datasets. The two frameworks learn to generate different types of errors, and taking advantage of both leads to substantial improvements in error detection.

We used the Approximate Randomisation Test \cite{Noreen1989,Cohen1995} to calculate statistical significance and found that the improvement for each of the systems using artificial data was significant over using only manual annotation. In addition, the final combination system is also significantly better compared to the \newcite{Felice2014a} system, on all three datasets.
While \newcite{Rei2016} also report separate experiments that achieve even higher performance, these models were trained on a considerably larger proprietary corpus.
In this paper we compare error detection frameworks trained on the same publicly available FCE dataset, thereby removing the confounding factor of dataset size and only focusing on the model architectures.

The error generation methods can generate alternative versions of the same input text -- the pattern-based method randomly samples the error locations, and the SMT system can provide an n-best list of alternative translations. Therefore, we also investigated the combination of multiple error-generated versions of the input files when training error detection models.
Figure \ref{fig:graph_increasing} shows the $F_{0.5}$ score on the development set, as the training data is increased by using more translations from the n-best list of the SMT system. 
These results reveal that allowing the model to see multiple alternative versions of the same file gives a distinct improvement -- showing the model both correct and incorrect variations of the same sentences likely assists in learning a discriminative model.
%While the benefits are much smaller, some performance improvements can also be observed from adding even more versions to the training data.
%We observed similar trends also using the pattern-based error generation method.

%The addition of either error generation system consistently improves performance on all benchmarks and the results are also statistically significant using the Approximate Randomisation Test \cite{Noreen1989,Cohen1995}.
%The combination of the original training set together with both pattern-based and SMT systems for error generation gives the best overall performance, improving $F_{0.5}$ by an absolute 5.39\% on the FCE test set. 
%The improvement of the combination is also significant over each of the individual systems on all three datasets. By giving the system access to data from different sources, generated by alternative approaches, the framework is able to learn the most robust error detection model.
%It should be noted that the results reported here are not directly comparable to the shared task results, as CoNLL-14 used a customised evaluation method and the participating systems were optimised using additional sources of training data.

\iffalse
\begin{figure}[t]
	\centering
	\includegraphics[width=0.8\linewidth]{graph_artdata_increasing.png}
	\caption{$F_{0.5}$ on FCE development set with increasing amounts of artificial data from SMT.}
	\label{fig:graph_increasing}
\end{figure}
\fi

\begin{figure}[t!]
		\centering

		\begin{tikzpicture}[scale=.75]
		\begin{axis}[axis x line=left, axis y line=left, xlabel=Number of versions, ylabel= $F_{0.5}$, legend style={at={(1,0.2)},
			anchor=north east, legend columns=3}, xmin=0, xmax=12, ymin=0.2, ymax=0.5] 
		%xmin=0, xmax=10, ymin=0, ymax=0.6, minor x tick num=1]
		\addplot+[] coordinates
		{(1,0.3915711188) (2,0.4459391965) (3,0.4615611635) (4,0.4676945084) (5,0.4705844367) (6,0.4633407007) (7,0.4679332335) (8,0.4750219772) (9,0.4739218572) (10,0.4767073419) (11,0.477776006)} 
		node[] {};
		\addlegendentry{FCE}
		
		\addplot+[] coordinates
		{(1,0.2735235463) (2,0.3925625896) (3,0.4039343121) (4,0.4146228343) (5,0.4133239364) (6,0.4104586499) (7,0.412831085) (8,0.4100146381) (9,0.4131427241) (10,0.4095500025) (11,0.4163970164)}  
		node[] {};
		\addlegendentry{EVP}
		
		\end{axis}
		\end{tikzpicture}
	\caption{$F_{0.5}$ on FCE development set with increasing amounts of artificial data from SMT.}
	\label{fig:graph_increasing}
\end{figure}
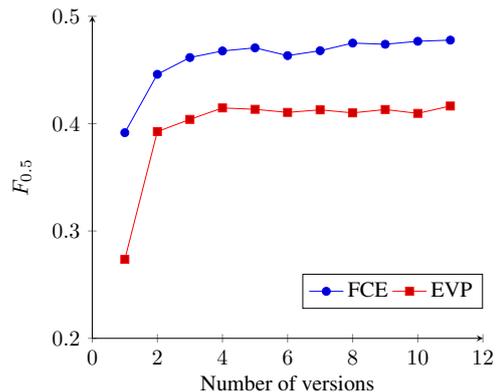

\section{Related Work}

Our work builds on prior research into AEG.
\newcite{Brockett2006} constructed regular expressions for transforming correct sentences to contain noun number errors.
\newcite{Rozovskaya2010a} learned confusion sets from an annotated corpus in order to generate preposition errors.
%\newcite{Rozovskaya2012} proposed an inflation method to boost error probabilities in order to increase recall. 
\newcite{Foster2009} devised a tool for generating errors for different types using patterns provided by the user or collected automatically from an annotated corpus. However, their method uses a limited number of edit operations and is thus unable to generate complex errors.
\newcite{Cahill2013} compared different training methodologies and showed that artificial errors helped correct prepositions.
\newcite{Felice2014a} learned error type distributions for generating five types of errors, and the system in Section \ref{sec:pat} is an extension of this model.
While previous work focused on generating a specific subset of error types, we explored two holistic approaches to AEG and showed that they are able to significantly improve error detection performance.

\section{Conclusion}

This paper investigated two AEG methods, in order to create additional training data for error detection.
First, we explored a method using textual patterns learned from an annotated corpus, which are used for inserting errors into correct input text.
In addition, we proposed formulating error generation as an MT framework, learning to translate from grammatically correct to incorrect sentences.

The addition of artificial data to the training process was evaluated on three error detection annotations, using the FCE and CoNLL 2014 datasets.
%The choice of input data was found to impact performance, with the FCE training data being best for experiments on FCE, whereas the Simple Wikipedia provided a good textual resource for the CoNLL dataset.
Making use of artificial data provided improvements for all data generation methods.
By relaxing the type restrictions and generating all types of errors, our pattern-based method consistently outperformed the system by \newcite{Felice2014a}.
The combination of the pattern-based method with the machine translation approach gave further substantial improvements and the best performance on all datasets.

%We found that both AEG methods were able to significantly improve error detection performance, with the SMT-based system achieving higher results compared to the pattern-based approach.
%Combining artificial errors from both methods was able to improve the performance further.

\bibliography{references,references-zheng}
\bibliographystyle{emnlp_natbib}

\end{document}